**Cognitive Anthropomorphism of AI: How Humans and Computers Classify Images**

Shane T. Mueller

Abstract/Feature at a glance:

Modern AI image classifiers have made impressive advances in recent years, but their performance often appears strange or violates expectations of users. This suggests humans engage in *cognitive anthropomorphism*: expecting AI to have the same nature as human intelligence.  This mismatch presents an obstacle to appropriate human-AI interaction. To delineate this mismatch, I examine known properties of human classification, in comparison to image classifier systems. Based on this examination, I offer three strategies for system design that can address the mismatch between human and AI classification:  explainable AI, novel methods for training users, and new algorithms that match human cognition.

In the past few years, enormous advances in artificial intelligence and machine learning have been made through the development of deep neural networks.  Some of the largest and earliest advances were in networks specializing in image labeling and classification, and still this domain is a large focus of research.  Along with exuberant enthusiasm for these systems, critics have responded by pointing out many of their limitations (Marcus, 2018).  One common criticism is that the systems are opaque and lack transparency, leading to the rebirth of the field of Explainable AI (XAI; see Swartout, 1977; Hoffman et al., 2017; 2018ab; Mueller, et al., 2018).  XAI systems have often attempted to combine existing AI systems, new algorithms for visualization and inference, and human-centered approaches to interaction design and interface in order to make systems that are understandable, predictable, transparent, and trustworthy.

   One hypothesis that may help designexplainable and transparent AIis a cognitive analog to the Computer are Social Actors (CASA) hypothesis (Nass & Moon, 2000). Nass and colleagues famously argued that humans treat computer systems as we do other people. But this may extend beyond social interaction--humans are also likely to anthropomorphize the *cognition* of AI systems by assuming it will operate in ways similar to human intelligence.  This approach to interacting with AI, which I call *Cognitive Anthropomorphism of AI,* is prevalent but frequently wrong: many capabilities of intelligent systems work in ways fundamentally different from similar human cognition.  Bos et al., (2019) introduced the similar concept of *effective anthropomorphism,* which argues that cognitive anthropomorphism can be an effective way to frame an initial mental model about AI.  However, it can also lead to mismatches and mis-predictions.  Although one hypothetical solution to these mismatches may be to mimic the biological algorithms in AI (see Samsonovich & Mueller, 2008; Mueller, et al., 2007), this is not the only approach.  If we recognize how human and machine cognition differ, this provides avenues for training, explanation, and future algorithm development that may help meet the needs and expectations of human users.

   In this commentary, I will focus on image classification, which is essentially a categorization problem, an area of cognition that has been extensively studied in humans for more than 50 years.  To understand what the human expectations might be, Table 1 summarizes many of the important major findings from the field that appear relevant to artificial image classification systems.  Each entry summarizes major domains of research, and each is meant as an entry point for interested readers to explore debates and theories within the human classification literature.  In addition, some of the important aspects are illustrated in Figure 1.

*Figure 1*. These silhouettes illustrate some of the systematic properties of human categories found in Table 1. The left column shows some examples of how human categorization is inferential: recognition is not generally impaired by occlusion of parts or non-accidental features, and we make inferences about the unseen properties even outside of the frame. Recognition is typically invariant to size, rotation, pose, and similar transformations—even the dog's shadow is recognizable (center top). Although many categories are similarity-based, they can also be based on theory (like the biological theory that unites different breeds of dogs that are often quite dissimilar in shape), or rules (e.g., pets are tame animals kept in the home). Categories have graded similarity among exemplars, but the exemplars may define the extent of the category (top right), although prototypical averages appear to used in some cases (center right), including allowing fast classification of some aspects before more details are available (global precedence). Finally, categories can be functional, based on use (bottom right).

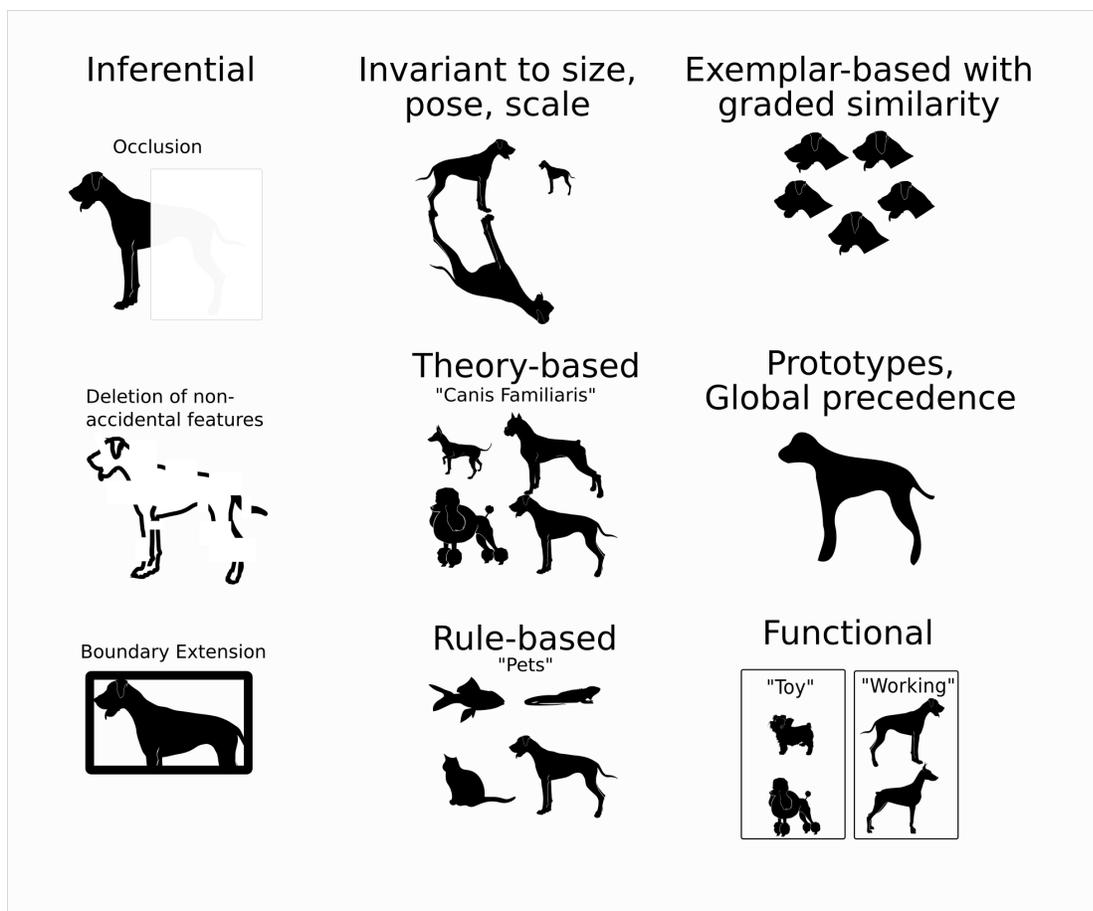

| Property | Reference | Description |
|---|---|---|
| **Representation** | | |
| a. Heterarchical | Collins & Loftus (1975) | Concepts are accessed in non-hierarchical ways. |
| b. Basic-level primacy | Rosch et al., (1976) | Basic level is entry point for human classification. |
| c. Prototype-based | Posner & Keele (1968); Smith & Minda (1998) | Categories can be defined by prototypical averages. |
| d. Exemplar-based | Nosofsky (1986); Kruschke (1992) | Categories can be defined by the exemplars. |
| | | |
| e. Variety of defining Principles | Murphy & Medin (1985); Smith & Sloman (1994); Gelman & Markman (1986) | Categories may be organized by rules, similarity, or theory. |
| f. Ad hoc | Barsalou (1983) | Ad hoc categories can be created without training. |
| g. Segmented Conceptually | Gelman (1988); Farah et al (1989); Sergent et al. (1992) | Qualitative differences in representation of, e.g. natural kinds vs. artifacts; faces and objects, etc. |
| h. Episodic | Standing (1973) | Accurate old-new classification of thousands of images suggests episodic (case-based) basis. |
| **Acquisition** | | |
| i. Learning | Homa et al (1973; 1981) | Abstract categories can be learned with minimal exemplars (3-9) and exposures (10-20). |
| j. Capacity | Standing et al. (1970), Standing (1973) | Recognition shows little loss for 2500-10,000 images. |
| | | |
| k. Inferential | Posner & Keele, (1968); Bransford & Franks (1971) | Unseen category prototypes are learned better than training exemplars; gist-based logical inferences are common. |
| l. Context-free | Grossberg (1994) | Object recognition is enabled by figure-ground separation. |
| m. Context-dependent | Palmer (1975); Oliva & Torralba (2007) | Classification is supported by context, especially with limited processing time. |
| n. Dependent on common ground | Klein, et al. (2005) | Classification serves a communication role that depends on shared understanding. |
| **Processing** | | |
| o. Global Precedence | Navon (1977) | Global shape properties is resolved before local features. |
| p. Shape-primacy | Pizlo (2010) | Object recognition proceeds via edge detection and shape reconstruction involving inference about 3D shape. |
| r. Functional | Martin et al. (1996); Chao & Martin (2000) | Visual objects are classified and represented functionally, rather than purely visually. |
| | | |
| s. What versus Where | Mishkin et al. (1983) | The dorsal and ventral visual pathways specialize respectively in spatial/relational and object processing. |
| **Invariants and Dependencies** | | |
| t. Retinal Scale | Larson (1985) | Size differences impair processing time moderately but have small impacts on accuracy. |
| u. Physical Scale | Rips (1989) | Classification can be dependent on physical scale. |
| v. Orientation/Pose | Tarr (1995) | Recognition of objects can depend on pose/viewpoint. |
| w. Occlusion | Biederman (1987); Pizlo (1994) | Recognition is robust to occlusion of non-defining features via inference of object properties. |
| x. Boundaries | Intraub & Richardson (1989) | Scene classification involves inference outside view. |

**Comparing Human and Machine Classification systems**

Cognitive scientists use a number of related terms to describe the processes that typically go on when an AI system processes an image and assigns a label.  The process of assigning a label is often called categorization or classification, which I will use interchangeably.  Most often, those labels refer to a set of things (a class) that we treat as equivalent; these are often referred to as *categories* or *classes*, or more generally *concepts*.  When other technical  terms are used, I will continue to give general definitions and examples of what is generally meant.

Table 1 illustrates many of the properties of human image classification.  I will discuss these properties in turn, using [] to reference specific entries in the table, and when appropriate compare human categorization to those typically exhibited by a number of a number of commercial and free deep image classification systems my colleagues and I have examined and tested alongside human participants (these including Google cloud vision client, IBM Watson Visual Recognition, Amazon Rekognition, Google Inception, and Clarifai General Image Classifier; cf. Hoffman, Klein & Mueller, 2018; Linja et al., 2019, for examples).

*Category-Level and hierarchy.* Researchers studying either human or AI categorization have recognized the need to understand category hierarchies.   These include subordinate/superordinate levels (lower versus higher levels, such as furniture→chair→office chair); part-whole relations, (hammer versus handle), and object-property relations (a diamond has the property hardness and clarity), and combinations of these (one might infer that a 'blue diamond' also has properties of hardness and clarity).

Currently, although most classifiers are trained on flat sets of category labels, some are trained on category hierarchies,  so that an image will be simultaneously classified and produce output on several levels (ideally, a picture of a pliers would be classified as a needlenose pliers, a pliers, a tool, and a product).  However, researchers have found that humans do not use strict hierarchical representations [a].  Rosch et al.'s (1976) finding  [b] indicate a middle-out preference for a *basic level of categorization:*  we typically prefer labels like "pliers" to either the superordinate "tool" or subordinate "needlenose pliers". In my laboratory, we have found that a major source of errors in machine image classifiers are these category-level errors (Linja et al., 2019).   This probably comes, in part, from communication norms that rely on common ground [n]-understanding shared knowledge and goals.  The label "pliers" is usually sufficient for communication, unless you need a specific kind to perform a particular task---if a master electrician asks an apprentice to "hand me a pliers", the apprentice might reasonably respond "which one".  Research has also suggested global-to-local processes [o] are prevalent during categorization, and so that we can categorize and resolve global properties (such as shape and size) before detailed local properties (e.g., details of a face) that help make fine distinctions between class members.

Modern image classifiers fail to address many of these category-level distinctions, for a variety of reasons.   Sometimes the failuresare based on the competitions and benchmarks used to develop and test the systems.  For example, many classifiers, including the popular VGG-16 model, are trained on 1000 categories defined as part of the ILSVRC Imagenet large scale visual recognition challenge[1]. This classification set  includes more than 100 types of dogs, and about as many kinds of birds,  but no "dog" or "bird" basic-level category. Thus, when given an unfamiliar dog it will incorrectly provide a specific dog category label, and the specificity of its answer (e.g., "Chesapeake Bay Retriever") may give the impression of a precision and confidence the system does not really have. But for other types of images (e.g., food),  the classes are at the basic level (e.g., pizza, bagel, potpie); the classifiers do not incorporate the subordinate-level categories such as 'pepperoni pizza', 'sesame seed bagel', or 'chicken pot pie', or the superordinate level category "food", both of which may be interesting.  Systems trained on category hierarchies produce results that are generally disappointing-- in our tests, a common label for a noisy picture of a hammer might be "product",  or  "metal"---both of which have elements of correctness but clearly wrong.

***Representation.***  Another long-standing debate in human classification involves assumptions about the fundamental representation of categories.  Researchers often think of a category as representing a class of things within an abstract semantic space, where there is a region of space whose members are typical of the category.  Different theories exist about how this category is represented, the simplest being as a prototype—something like the average or most typical example [c].  A category might be represented efficiently as the center of the set of examples [c], perhaps with the variability along different axes, but this will be inadequate if the shape of the category is irregular, which has led others to suggest the category is maintained less efficiently as a set of examples (or exemplars) [d].  The debate is still ongoing, with the greatest advances coming from computational models that implement and test assumptions about human category representation.  Insofar as neural networks work by error-reduction, this may appear to favor prototype representations.  However, Kruschke (1991) demonstrated that exemplar-based neural networks are reasonable and highly capable, and visualization of features spaces within deep networks appears to show representations of a range of features typical of the instances of a category (e.g., Zeiler & Fergus, 2014), perhaps suggesting artificial neural networks are somewhat exemplar-based.

Exemplar approaches are good at representing categories that have many highly-distinct members (e.g., chairs), or have properties that are often correlated (e.g., leg length and neck length in quadroped mammals) or are within a space that their average does not describe any example well (e.g., the minerals studied by Nosofsky et al., 2017).  Indeed, some of the systematic failures of image recognition networks may occur to the extent that the AI learns an average prototype representation, while humans  are able to segregate them into multiple exemplars.

***Defining Properties.*** Categorization research has examined the defining properties of human categories [e].  Although classic theories of categorization were typically based on the similarity of the features of category members to one another or to a prototype (all Brittany spaniels share a common size, shape, and disposition), but there are many examples where a category can be defined by a rule (a hunting dog is one that points or retrieves), or even more abstract theories (a Brittany was bred as a cross of English pointers and French hunting dogs). Furthermore, research also suggests human categories have graded or fuzzy structures. This means that (1) there are cases that are more typical of the category than others; (2) there are unclear cases; and (3) there are non-category members that vary in their similarity to the category. Properties are *characteristic* rather than *defining,* allowing for exceptions and ambiguities. Because most image classifiers are trained with human-generated categories, they will tend to learn responses consistent with these categories, but they may be just as easily trained on categories that violate these properties of human categories. Moreover, categories based on rules and theories (classifying a dolphin as a mammal rather than a fish) are likely to cause generalization problems unless trained specifically to avoid them.

Categories can also be [f] *ad hoc:* created on the spot, rather than through training (an example is "*things to take on a camping trip*"). These categories show the same graded structure as other human categories, but they are unlike the categories learned by image classifiers (which require extensive training on preset classes). Ad hoc categories require inference and search through existing knowledge, which may require mechanisms that typically do not exist in current image classifiers. This suggests human categories are, at least to some extent, created as needed to suit the situation, which is very different from how image classifiers learn categories. Human categories can also be based on the *function* of members rather than visual properties [g], even to the extent that, for example, neural representations of tools activate motor regions of the cortex that capture how the tools are used.

So, with respect to these defining properties of categories, image classification categories appear quite limited; and resemblance to human categories is as likely as not smuggled in through human-designed training sets rather than underlying algorithms or architectures. Certainly, AI categories exhibit characteristic properties and graded structure, but this is partly because the classes they are trained on exhibit these properties already. Consequently, AI categorization is currently good only at limited types of categories important for human use. Yet users may not understand this, and expect AI systems to work with human-level categories, which offers challenges for designing the interaction of the systems.

***Brain Processes and Representations.*** Categorization has also been studied by cognitive neuroscientists, who examine the specialized brain structures involved in the categorization process. Human categorization is subject to very real computational constraints created at evolutionary timescales, and may not solely be learned from the statistical structure of the world. For example, researchers have distinguished between brain networks used for rule versus similarity-based categories (e.g., Patalano, et al., 2001); identified neural representations of functional categories [r]; and delineated many well-understood neural processes in visual and object categorization. This starts with processing in the retina, moves through the visual cortex [t,u,v], and then forward in the brain via parallel pathways specializing in object and spatial information (what-versus-where pathways [s]), until merging again in frontal regions, at which point different classes of objects are sometimes represented in different regions of the brain [g,r]. Each of these steps leave characteristic fingerprints regarding how human categorization occurs, and the specialized processing occurring in different cortical regions (e.g., reasoning involved in prefrontal cortex, function involving motor areas) may require similar specialization in artificial systems.

Current classification networks do handle some of these processes—the convolutional neural network architecture in particular enables recognition regardless of position, and thus supports the separate processing of what-versus-where information. However, the specialized processing of different brain networks may help inform new artificial architectures that have similar specialized representations. Furthermore, aspects of this knowledge can be helpful for designing human-AI interfaces. Many errors in image classifier systems appear to stem from these sources, including the very notion of what an object is, which is a concepts for which AI systems often have no clear representation.

**Learning.** In general, learning in artificial neural networks is not only slow, it requires many examples to enable robust performance. In contrast, humans can learn new categories with minimal exposures and examples [i]. It is difficult to determine whether the networks encode specific example images, but there is a sense in which people do [h]: people are able to encode and recognize many thousands of images after just one exposure. This large capacity [j] is remarkable, and typically not a capability of image classification networks. The AI community recognizes that requirements for large amounts of data is problematic, and some researchers have made efforts to improve visual classification from relatively few training samples (e.g., Pascuale et al., 2016). This difference between humans and AI may reflect aspects of inference and processing that make human categorization fundamentally different from machine classification systems, and these differences could be especially important for users to recognize. For example, an AI system may fail to correctly classify an image that it was trained on, or may fail in a new domain because it was not trained with sufficient examples. Users may not understand why it fails in either of these cases, but appropriate user-centered design may help by communicating how confident or robust the system's answers are likely to be.

***Visual Properties and Invariants.*** Many properties of human visual categorization have been learned by discovering *invariant* properties: aspects of the system that **are not impacted** by some change. For example, line-drawing caricatures simplify representations of faces, and accentuate differences from the average. Because this is consistent with the processes involved in human facial recognition, caricatures generally don't impair and can make recognition easier than a photograph (Mauro & Kubovy, 1992). Thus, human facial recognition is invariant to the steps involved in making caricatures, and this may reveal that human face recognition works in ways similar to caricature drawing. Similarly, line drawing contours of an object generally do not impair human recognition, nor do rotations or context changes (moving an object to a background it does not usually appear in), because they retain the critical features used in object recognition, while also removing many of the features that are irrelevant (Pizlo, 2010). Yet these all impact image classifiers. Many of these invariants appear to be the consequence of how human object classification depends critically on representations of 3D shape--relying on assumptions about the object's symmetry or compactness (Pizlo, 2010).

Machine classification systems that do not make inferences about 3D shape are likely to suffer difficulty with many of the visual invariants of human object recognition.

Human classification is often invariant to many low-level properties of objects as well, including scale [t,u], pose and or orientation [v], occlusion [w], boundaries [x], and many transforms of shape [p]. As discussed earlier, the visual system separates spatial and object information along separate pathways [s], so that position is explicitly decoupled by brain specialization. But these invariants are not universal and depend on context [m,u]: physical scale is not always ignored (as when comparing a real versus model racecar). Indeed, context itself has opposing influences on human categorization: classification requires segmenting a target from its background context so that classification can be made [l], but the context or background can help or hurt the speed and ease of the classification [m]. Furthermore, because real-world scale is rarely present in imagery, most current image classifier systems have little or no ability to use or make inferences about it. There are also non-visual aspects of categorization that may be considered invariants. For example, memories and stories are stored by gist and inference, so transformations of the surface form that preserve the meaning are not detected as new information [k].

Altogether, this knowledge can be useful to designers of AI systems because it may help provide better guidance to users about how to best use the system. Humans may not recognize all the ways in which an image classifier might be impaired but human classification is spared. Recognizing this, training and instruction may help improve the use of a system (for example, to use an app that searches for a certain product based on a picture taken by a phone, the user could be instructed picture should be taken on a solid background, and in a standard orientation, in bright light, etc.)

**Implications for the design of human interaction with intelligent systems**

So far, this paper has described a hypothesis that human users will assume AI cognition is human-like (cognitive anthropomorphism), and examined some of the most broadly-understood phenomena and theories within human classification to articulate (1) how human classification works, and (2) how machine classification typically differs from human classification. This establishes potential areas of mismatch between human expectation and AI capability. There are three general design strategies that can be used to address this mismatch:by changing the AI, the human user, or the interaction between human and AI.

***Change the AI***. One of the design consequences of Nass's CASA thesis was to embrace *social anthropomorphism*, and build computer systems that behaved as social actors. Although this approach has been derided because of some prominent mis-steps (e.g., Microsoft Bob; Clippy) there are strategies for AI design that may help by changing the system to behave as naive users might expect. One strategy involves biomimicry or biological inspiration (cf. Samsonovich & Mueller, 2008)— attempts to generate AI systems that succeed and fail in the same ways humans do. This approach has not been proven to be a fix-all, and it may be aiming for the wrong outcome. The artificial intelligence systems we use and rely are are adopted because they perform tasks better, faster, or more cheaply than humans can, and so it may not be productive to hobble them, for example, by inducing behaviors that humans exhibit to their detriment (e.g., base rate neglect, decision biases, forgetting, etc.).

Another approach is to improve and adapt AI architectures, algorithms, and training sets to improve the performance with respect to human expectation, while at the same time improving or having no negative impact on the system capability .  Some central properties of human classification might be produced by current image classifiers using new training sets, or relatively simple reorganization of current network systems, but others may require substantial development of new algorithms or adoption of other logic not used in deep learning networks. For example, shape-based classification algorithms are able to account for many aspects of how humans use 3D shape and contour to recognize objects; these are areas that are especially challenging for deep learning networks, and developing ways to integrate shape-base representations into image classifiers may create a substantial advances.

***Change the human.*** A second approach that has been used in the past, and is likely to continue to be used, is to develop training and instruction that informs human expectations about the AI (see Schniederman, et al., 2016, Chapter 14).  Mueller & Klein (2011) proposed methods for training humans to understand intelligent systems in the form of an 'experiential user guide', and Mueller & Tan (2019) similarly discussed design of a cognitive tutorial for improving understanding.  These approaches rely on (1) identifying the mismatch between human expectation and AI behavior, especially in the form of the boundary conditions of performance and expectation and (2) providing explicit example-based training on these boundaries, to help users anticipate and work around unexpected performance.  The design of this kind of training differs from typical approaches that primarily focus on the capabilities of a system, because it highlights both how the system works and how it fails. For some of the properties of human classification that differ from AI, instruction about expectations may be the best approach.  For example, global precedence of object recognition means that humans may detect general properties in an image or video (i.e., that a form is a person) even if details cannot be detected (i.e., gender or age).  But blurring can often severely impair performance of an AI system. A user who is trained to understand this limitation will know the contexts in which the system can be relied on, and those that it should not be.

***Change the interaction.*** A third general approach is to focus on ways to change and improve the interaction between human and machine.  The XAI approach is maybe the best example of this, because these typically find ways to expose the inner workings of the AI to the user, but without change the AI algorithms themselves.  Work on XAI has often involved visualization algorithms, such as Ribiero et al's (2016) LIME, which provide interfaces to visualize the importance of features used to make classification decisions.  These visualization algorithms alone are unlikely to be sufficient generally, and more serious interaction design is necessary to understand the shared human-AI tasks and goals, and to determine if the algorithm helps expose information that can combat a misunderstanding of the system.  One promising approach is to develop interactions that support sensemaking and self-explanation of the user--identifying information and interfaces that enable exploring, comparing, contrasting, and testing differences in stimuli and classifications to allow a user to generate a robust mental model of the process.

**Summary and Conclusions**

Overall, human classification differs from AI systems in many ways. Humans using these system may incorrectly assume they operate like humans do, which I refer to as the cognitive anthropomorphism of AI. This mismatch means that humans may reject technology that is useful, unless systems are appropriately designed to acknowledge the mismatch and work against it. This can be done in a number of ways, but all involve some level of interactive design in order to improve human-machine capabilities. Thus, the comparison between human and machine classification described here may offer a number of leverage points to improve AI-human capabilities.

**Acknowledgments**

This work was supported by DARPA's Explainable AI Program. I thank Robert Hoffman, Gary Klein, and Tim Miller for supportive commentary and guidance. Imagery in Figure 1 was adapted from CC-Attribution 4.0 licensed content at https://www.freevector.com/pet-animals and https://www.freevector.com/dogs.

**Biography**
Shane T. Mueller is an Associate Professor in the Department of Cognitive and Learning Sciences at Michigan Technological University.   He specializes in computational modeling of human performance, knowledge, and memory phenomena, in basic and applied contexts.  His research interests involve Explainable AI, models of human cognitive and performance impact from environmental stressors, expertise in word game enthusiasts including high-performance crossword players, and he is the developer of the Psychology Experiment Building Language (PEBL), a computerized framework for psychological testing.

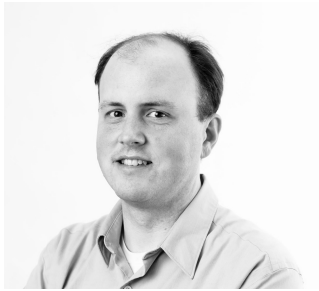